\documentclass[runningheads]{llncs}

\usepackage{epsfig}
\usepackage{subcaption}
\usepackage{calc}
\usepackage{amssymb, amsfonts}
\usepackage{amstext}
\usepackage{amsmath}
\usepackage{multicol}
\usepackage{multirow}
\usepackage{pslatex}
\usepackage[linesnumbered,lined,ruled]{algorithm2e}
\usepackage{enumitem}
\usepackage[bottom]{footmisc}
\usepackage{url}
\usepackage{booktabs}
\usepackage{floatrow}
\usepackage{bbding}
\usepackage{pifont}
\usepackage{wasysym}

\usepackage[pagebackref=false,breaklinks=true,letterpaper=true,colorlinks,bookmarks=false]{hyperref}

\usepackage[backend=biber,style=numeric,sorting=none]{biblatex}
\begin{document}

\title{Enabling On-device Continual Learning \\ with Binary Neural Networks}

\author{Lorenzo Vorabbi\inst{1,2}\orcidID{0000-0002-4634-2044} \and
Davide Maltoni\inst{2}\orcidID{0000-0002-6329-6756} \and
Guido Borghi\inst{2}\orcidID{0000-0003-2441-7524} \and
Stefano Santi\inst{1}}
\authorrunning{L. Vorabbi et al.}
\institute{Datalogic Labs, Bologna 40012, IT \and
University of Bologna, DISI, Cesena Campus, Cesena 47521, IT
\email{\{lorenzo.vorabbi2,davide.maltoni\}@unibo.it}\\
\email{\{lorenzo.vorabbi,stefano.santi\}@datalogic.com}}

\maketitle

\abstract{On-device learning remains a formidable challenge, especially when dealing with resource-constrained devices that have limited computational capabilities. This challenge is primarily rooted in two key issues: first, the memory available on embedded devices is typically insufficient to accommodate the memory-intensive back-propagation algorithm, which often relies on floating-point precision. Second, the development of learning algorithms on models with extreme quantization levels, such as Binary Neural Networks (BNNs), is critical due to the drastic reduction in bit representation. In this study, we propose a solution that combines recent advancements in the field of Continual Learning (CL) and Binary Neural Networks to enable on-device training while maintaining competitive performance. Specifically, our approach leverages binary latent replay (LR) activations and a novel quantization scheme that significantly reduces the number of bits required for gradient computation. The experimental validation demonstrates a significant accuracy improvement in combination with a noticeable reduction in memory requirement, confirming the suitability of our approach in expanding the practical applications of deep learning in real-world scenarios.}


%\onecolumn \maketitle \normalsize \setcounter{footnote}{0} \vfill
\setcounter{footnote}{0}

\section{Introduction}
\label{sec:introduction}

\begin{figure*}[!h]
\centering
\includegraphics[width=1.0\linewidth,scale=0.8]{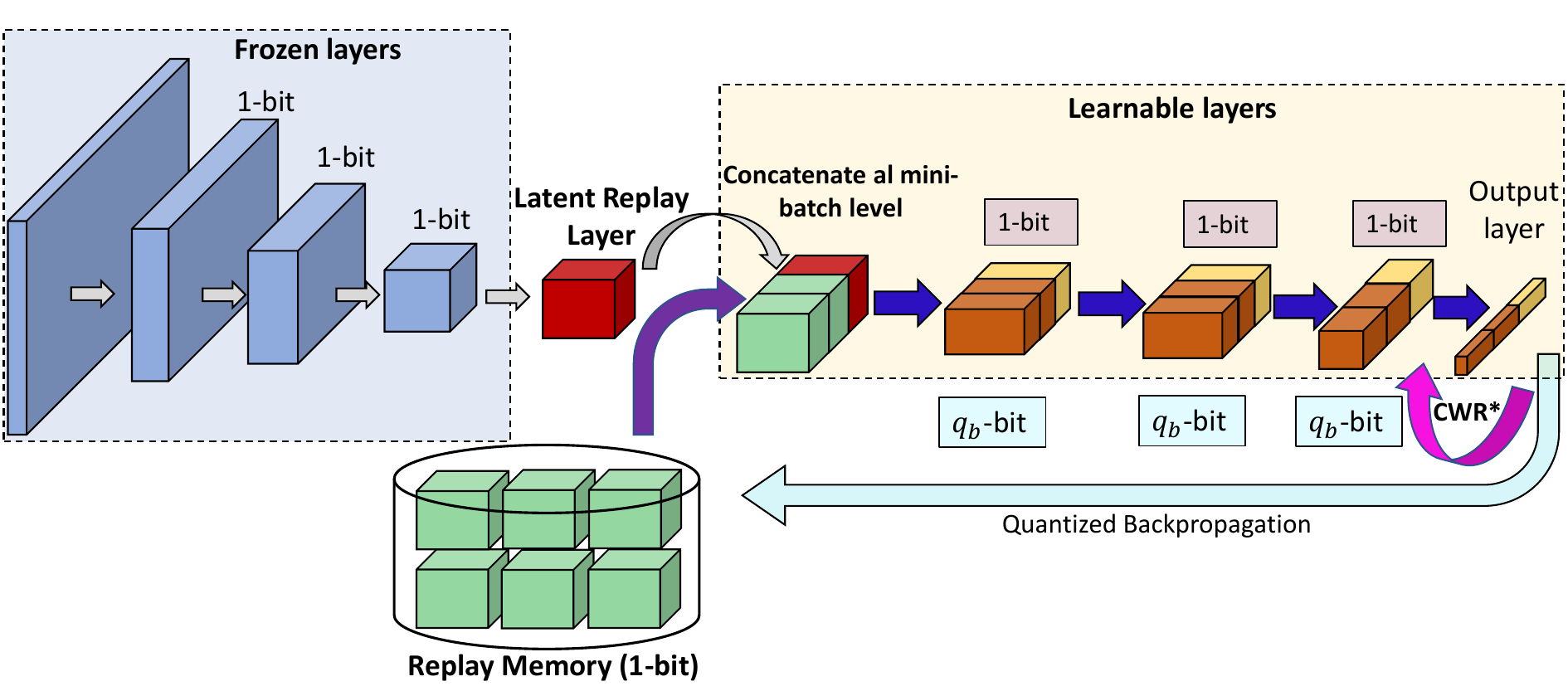}
  	\caption{Continual Learning with latent replay memory. When using a BNN the activations stored in the replay memory can be quantized to 1-bit.}
  	\label{fig:latent_replay_schema}
\end{figure*}

In recent times, the integration of Artificial Intelligence into the Internet of Things (IoT) paradigm~\cite{mohamed2020relation, alshehri2020comprehensive}, enabling the provision of intelligent systems capable of learning even within embedded or tiny devices, has garnered significant attention in the literature. 
This trend has been facilitated by various factors, including the evolution of microchips, which have led to the availability of cost-effective chips in many everyday objects. Additionally, the exploration of new learning paradigms, such as Continual Learning (CL)~\cite{parisi2019continual, masana2022class}, has contributed to the development of techniques for training neural networks continuously, on small data portions (denoted as \textit{experiences}) at a time, mitigating the issue of catastrophic forgetting~\cite{kirkpatrick2017overcoming}. In this manner, a neural network, in contrast to the traditional machine learning paradigm, does not learn from a single large dataset accessible entirely during the training phase but rather from small data portions accessible gradually over time.
This limited amount of data needed by the training procedure effectively simplifies the adoption of a CL training implementation on embedded devices. 

Despite the keen interest of the scientific community, numerous challenges still persist, rendering the utilization of deep learning models on devices particularly demanding. These challenges are primarily associated with the computational requirements typically demanded by deep neural networks, even though based on CL strategies. Indeed, embedded devices often have limited available memory, preventing the storage of a vast amount of data. Furthermore, a powerful GPU is usually absent due to cost, space constraints, and energy consumption. These competing needs have given rise in the last few years to a specific branch of machine learning and deep learning called TinyML~\cite{banbury2020benchmarking}, focused on shrinking and compressing neural network models with respect to the target device characteristics. One of the most interesting TinyML approaches, is Binary Neural Networks (BNNs)~\cite{courbariaux2016binarized, rastegari2016xnor, qin2020binary}, where a single bit is used to encode weights and activations; unfortunately, solutions based on BNNs in combination with Continual Learning algorithms are still lacking.

A previous work~\cite{vorabbi2023device} explored the possibility of training a BNN model on-device by freezing the binary backbone and allowing the adaptation of only the last classification layer where forgetting is mitigated by CWR*~\cite{lomonaco2020rehearsal, graffieti2022continual}. Unfortunately, the reported results are interesting but the final accuracy is significantly lower w.r.t. a system where all the layers can be tuned.
Pellegrini et al.~\cite{pellegrini2020latent} showed that a good accuracy/efficiency tradeoff in CL can be achieved by only some convolutional layers (typically from $3$ to $5$), placed before the classification head.
Replaying part of old data (stored in a replay memory or buffer), interleaved with new samples, was proved to be an effective approach to mitigate catastrophic forgetting~\cite{kirkpatrick2017overcoming}. If past samples are stored as intermediate activations (instead of raw data), the replay technique takes the name \textit{latent replay}~\cite{pellegrini2020latent} (see Fig. \ref{fig:latent_replay_schema}). Latent replay is particularly interesting when combined with BNN (as proposed in this paper) since the latent activations can be quantized to 1-bit, leading to a remarkable storage saving. Unfreezing some intermediate layers requires to back-propagate gradients along the model to update weights; on the edge, the implementation of this process, usually referred to as \textit{on-device} learning, requires an efficient and lightweight back-propagation implementation, which is not yet available in the most popular training frameworks. The reduction of bitwidths during backward pass, made possible by a fixed point (many low-power CPUs are not equipped with floating-point unit) implementation, can speedup the learning phase but the tradeoffs between accuracy loss and efficiency need to be evaluated with attention.

In this paper, we propose a solution to combine the Continual Learning paradigm with training on the edge using BNNs. Specifically, through the introduction of a back-propagation and input binarization algorithm, we demonstrate how it is possible to continuously tune a CNN model (including classification head and convolutional layers) with low memory requirements and high efficiency. Our work represents a step beyond the classical quantization approach of BNNs published in the literature, where binarization is typically considered only during forward pass and a binary model is trained using latent floating-point weights~\cite{helwegen2019latent}. Some works showed~\cite{cai2020tinytl, lin2022device} good improvements in reducing both the memory demand and the computational effort to enable training on the edge, but they did not focus on the Continual Learning (CL) scenario, which we primarily address.
We conducted experiments with multiple BNN models, evaluating the advantages offered by the proposed methodology in comparison to the method outlined in~\cite{vorabbi2023device} where only the classification head is tuned.  

The main contributions of this work can be summarized as follows:

\begin{enumerate}
    \item \textbf{Reduced Replay Memory Requirement}: our replay memory stores intermediate activations quantized to 1-bit allowing a relevant storage saving. We investigate the trade-offs required to maintain model accuracy while simultaneously reducing memory consumption.
    \item \textbf{Improved Model Accuracy}: by enabling the continual adaptation of intermediate convolutional layers (besides the final classification head) our BNN-based model significantly outperforms the closest previous solution ~\cite{vorabbi2023device}.
    \item \textbf{Quantization of Backpropagation for Non-Binary layers}: we introduce a quantization approach for the back-propagation step in non-binary layers, enabling the preservation of accuracy while eliminating floating-point operations.
    \item \textbf{Optimized Binary Weight Quantization}: we present an optimized quantization strategy tailored for binary weights, leading to a remarkable $8\times$ reduction in memory requirements. Binary layers are typically trained by storing latent floating-point representations of weights that are subsequently binarized during inference. Replicating this schema on-device would result in an unacceptable increase of memory usage and computational overhead.
    \item \textbf{Optimized Back-Propagation Framework}: we implemented a comprehensive back-propagation framework capable of supporting various quantization levels both inference and back-propagation stages.
\end{enumerate}

The paper is organized as follows.  
In Section \ref{latent_replay_sec} we describe the latent replay mechanism providing an estimation of the memory saved when applied to a binary layer. 
In Section \ref{quant_strategy} we detail the quantization approach used for both forward and backward passes. 
Then, in Section \ref{quantized_back_prop} we describe the method used to quantize gradient computation.
A comprehensive experimental evaluation is proposed in Section \ref{section_experiments}, focusing on the accuracy comparison with respect to the CWR* algorithm (Sect. \ref{section_accuracy_result}), the reduction of the storage needed by the replay memory (Sect. \ref{reduction_memory}) and the efficiency in the backpropagation algorithm (Sect. \ref{eficiency_estimation}).

\section{Related Work}
\label{related_work}

\begin{description}
\item{\bfseries{Continual Learning}}: Some works in the literature addressed the on-device learning task proposing solutions to primarily reduce the memory requirement of the learning algorithm: Ren et al.~\cite{ren2021tinyol} brought the transfer learning task on tiny devices by adding a trainable layer on top of a frozen inference model. Cai et al.~\cite{cai2020tinytl} proposed to freeze the model weights and retrain only the biases reducing the memory storage during forward pass. Lin et al.~\cite{lin2022device} introduced a sparse update technique to skip the gradient computation of less important layers and sub-tensors. QLR-CL~\cite{ravaglia2021tinyml} relies on low-bitwidth quantization (8-bit) to speed up the execution of the network up to the latent layer and at the same time reduce the memory requirement of the latent replay vectors from the 32-bit floating point to 8-bit; compared to our solution, QLR-CL optimizes the computation pipeline for a specific ultra low-power CPU based on RISC-V ISA. In addition, backpropagation is performed with floating-point precision. In \cite{nadalini2022pulp, nadalini2023reduced}, Nadalini et al. introduced a framework to execute on-device learning on tiny devices using floating-point (32 and 16 bits) computation. Our solution differs considerably even in this case because we introduce a quantized fixed-point implementation for binary and non-binary layers instead of performing a post-training quantization of the frozen layers and then executing backpropagation with floating-point precision. Additionally, to the best of the author's knowledge, this is the first work that implements on-device learning by quantizing the back-prop of binary layers using low bitwidths.

As in \cite{vorabbi2023device} the proposed approach uses CWR* for class-bias correction in the classification head (see \cite{masana2022class}). CWR* maintains two sets of weights for the output classification layer: $cw$ are the consolidated weights used during inference while $tw$ are the temporary weights that are iteratively updated during back-propagation. $cw$ are initialized to $0$ before the first batch and then updated according to Algorithm 1 of \cite{lomonaco2020rehearsal}, while $tw$ are reset to $0$ before each training mini-batch. CWR*, for each already encountered class (of the current training batch), reloads the consolidated weights $cw$ at the beginning of each training batch and, during the consolidation step, adopts a weighted sum based on the number of the training samples encountered in the past batches and those of current batch.

\item{\bfseries{Binary Neural Networks}}: Quantization is a useful technique to compress Neural Network models compared to their floating-point counterparts, by representing the network weights and activations with very low precision. The most extreme quantization is binarization, where data can only have two possible values, namely $-1\left(0 \right)$ or $+1\left(1 \right)$. By representing weights and activations using only 1-bit, the resulting memory footprint of the model is dramatically reduced and the heavy matrix multiplication operations can be replaced with light-weighted bitwise XNOR operations and Bitcount operations. According to \cite{bannink2021larq}, which compared the speedups of binary layers w.r.t. the 8-bit quantized and floating point  layers, a binary implementation can achieve a lower inference time from $9$ to $12 \times$ on a low power ARM CPU. Therefore, Binary Neural Networks combine many hardware-friendly properties including memory saving, power efficiency and significant acceleration; for some network topologies, BNN can be executed on-device without the usage of floating-point operations~\cite{vorabbi2023optimizing} simplifying the deployment on ASIC or FPGA hardware.
\end{description}

\begin{figure*}[!h]
\centering
\includegraphics[width=1.0\linewidth]{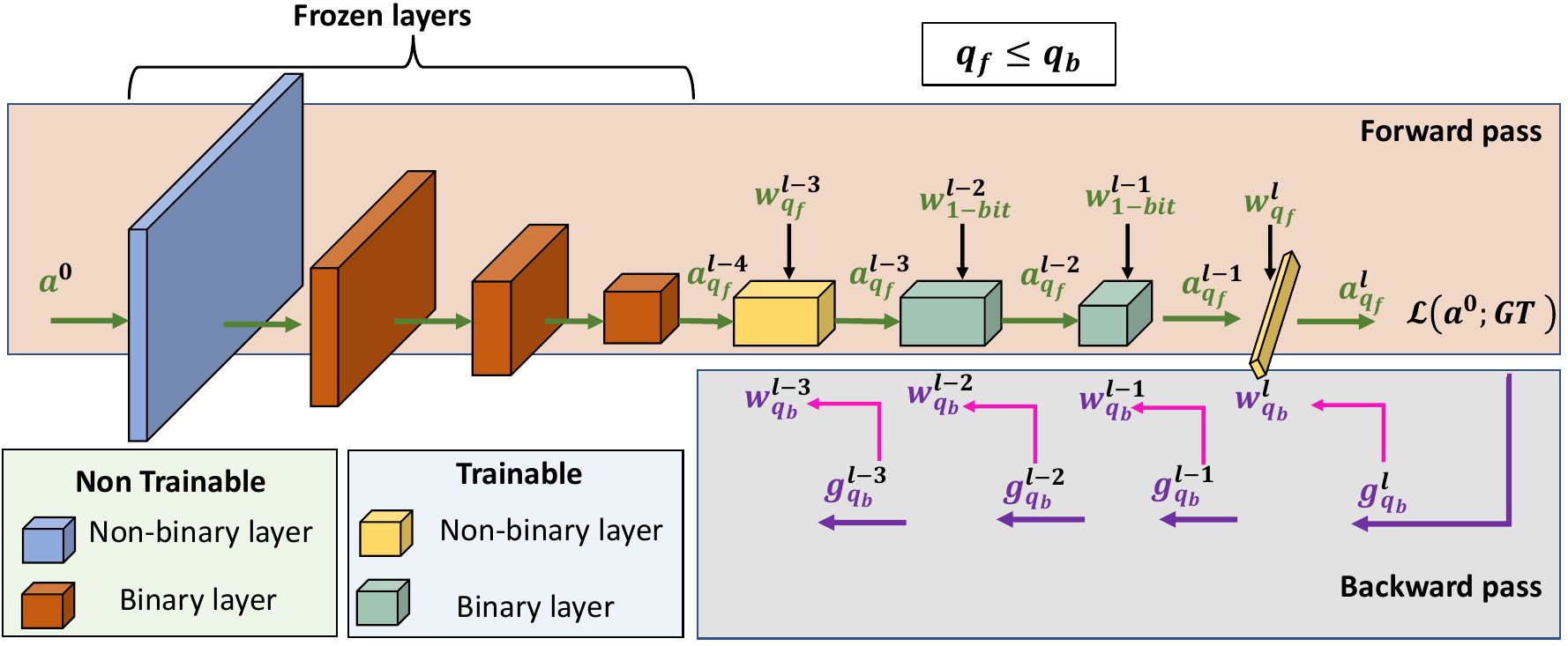}
  	\caption{Quantization scheme that uses a different number of bitwidth for forward ($q_{f}$) and backward ($q_{b}$) pass. Usually, trainable non-binary layers are Batch Normalization~\cite{ioffe2015batch}, Addition and Concatenation layers.}
  	\label{fig:quantization_scheme}
\end{figure*}

\section{Method}
In this section, we introduce our solution to efficiently deploy CL methods using Latent Replay and BNNs. In particular, the CWR* approach (briefly summarized in Sect. \ref{related_work}) is used to correct class-bias in the classification head.
%, but this remains unchanged with respect to \cite{vorabbi2023device} and the reader can find more details in the paper.%

\subsection{Continual Learning with Latent Replays}
\label{latent_replay_sec}
In Fig. \ref{fig:latent_replay_schema} we illustrate the CL process with Latent Replay. 
When new data becomes available, they are fed to the neural network that during the forward pass produces their latent activations, which represent the feature maps corresponding to a specific intermediate layer. We denote this layer as $l$ (where $l \in \left [0, L  \right )$), with $L$ representing the total number of layers within the model. Activations of new data are joined (at minibatch level) with the replay activations (previously stored) and forward/backward passes on the remaining layers, specifically those with index from $l+1$ to $L-1$. To elucidate further, if $B_{N}$ denotes the minibatch size of the newly acquired latent activations, a subset of replay vectors ($B_{R}$) is extracted from the replay memory and merged, thus forming a minibatch of total size $B_{T}=B_{N}+B_{R}$. 
In contrast, the layers with an index less than $l$ are maintained in a frozen state and are not included in the learning process.
After the conclusion of each training experience, the replay memory is updated by including samples from the last experience and using class-balanced reservoir sampling~\cite{vitter1985random}, which ensures a double balancing: (i) in terms of samples per classes, (ii) in terms of samples from experience (see Algorithm \ref{algorithm_replay_memory}).

\begin{algorithm}[!h]
 \caption{Procedure used to populate the replay memory (RM). RM is initially pre-populated using training samples of the first experience. The reservoir sampling is used on a class basis to maintain the balance among different classes. This approach prevents a skewed representation of classes within RM. }
 \label{algorithm_replay_memory}
 \KwIn{$N=$ max number of samples per class}
 \KwIn{$C=$ max number of classes}
 $RM_{size}=C\cdot N$\; \tcp{$C\cdot N$ is the max size of RM populated during the first experience.}
 \BlankLine
 \For{each on-device experience}{
    \tcp{$T$ is the number of classes}
    \tcp{$M_{t} =$ samples of class $t$}
    \tcp{$RM_{t} =$ samples of class $t$ already in RM}
    \For{$t=0$ \KwTo $T-1$}
    {$B_{t} = RM_{t} \cup  M_{t}$ \;
      \tcp{$\#$ is the cardinality operator}
      $RM_{t}^{new}=$ apply Reservoir sampling to extract $\#RM_{t}$ samples from $B_{t}$\;
      remove not selected $RM_{t}$ samples \;
      update RM with $RM_{t}^{new}$\;}
  }
\end{algorithm}

\subsection{Quantization of activations and weights}
\label{quant_strategy}
Quantization techniques have gained widespread adoption to diminish the data size associated with model parameters and the activations of layers. Employing quantization strategies enables the reduction of data bitwidth from the conventional 32-bit floating-point representation to a lower bit-precision format, typically 8 bits or even less, while typically incurring a negligible loss in accuracy during the forward pass of the model. For the quantization of non-binary layers that need to be trained on-device, we adopted the approach proposed in the work of Jacob et al.~\cite{jacob2018quantization} which is further implemented in the GEMMLOWP library~\cite{jacob2017gemmlowp}.

By representing the dynamic range of the activations at the $i$-th  layer of the network as $\left [a_{min}^{i}, a_{max}^{i}  \right ]$, we can define the quantized activations $a_{q}$ as:

\begin{equation}
\label{quant_formula}
\begin{matrix}
 a_{q}^{i} = cast \textunderscore to \textunderscore q\lfloor \frac{a^{i}}{S_{a}^{i}}  \rceil,
 &
 S_{a}^{i} = \frac{a_{max}^{i} - a_{min}^{i}}{2^{q}-1}
\end{matrix}
\end{equation}

where $q$ denotes the number of quantization bits used $\left ( 8, 16, 32 \right )$, $a^{i}$ represents the full-precision activations and $a_{max}^{i}$, $a_{min}^{i}$ are determinated through calibration on the training dataset. Weight quantization can be accomplished using an equation analogous to equation \ref{quant_formula}.
However, as recommended in Vorabbi et al.~\cite{vorabbi2023device}, we utilize two separate sets of quantization bits for both the forward and backward passes.
For binary layers, during the forward pass, binarization is executed according to the following equation:

\begin{equation}
\label{set_eq}
    STE \left( x\right) = \left\{\begin{matrix}
+1 & x\geq 0 \\ 
-1 & otherwise
\end{matrix}\right.
\end{equation}

as proposed in \cite{courbariaux2016binarized}. In backward pass, STE computes the derivative of sign as if the binary operation was a linear function. This approximation has been further improved by other works~\cite{liu2018bi, liu2020reactnet} and in general it is model dependent.

\begin{table*}[!t]
\centering
\begin{tabular}{c|c|c|c|c|cl}
\toprule
  & \# total weights & LR shape & \begin{tabular}{@{}c@{}} \# $B=$ binary \\ weights after LR\end{tabular} & \begin{tabular}{@{}c@{}} \# $NB=$ non-binary \\ weights after LR\end{tabular} & $\frac{B}{B+NB} $ & \\
 \midrule
 \textit{BiReal-18} & $11.2M$ & $\left ( 4, 4, 512 \right )$ & $7.0M$ & $19K$ & $99.7\%$ & \\
 \textit{BiReal-18} & $11.2M$ & $\left ( 8, 8, 256 \right )$ & $10.1M$ & $28K$ & $99.7\%$ & \\
 \textit{React-18} & $11.1M$ & $\left ( 4, 4, 512 \right )$ & $7.0M$ & $18K$ & $99.7\%$ & \\
 \textit{React-18} & $11.1M$ & $\left ( 8, 8, 256 \right )$ & $8.3M$ & $24K$ & $99.7\%$ & \\
 \textit{VGG-Small} & $4.6M$ & $\left ( 8, 8, 512 \right )$ & $2.3M$ & $86K$ & $96.4\%$ & \\
 \textit{QuickNet} & $12.7M$ & $\left ( 7, 7, 256 \right )$ & $9.5M$ & $36K$ & $99.6\%$ & \\
 \textit{QuickNet} & $12.7M$ & $\left ( 14, 14, 128 \right )$ & $11.9M$ & $43K$ & $99.6\%$ & \\
 \textit{QuickNetLarge} & $22.8M$ & $\left ( 7, 7, 256 \right )$ & $14.2M$ & $40K$ & $99.7\%$ & \\
 \textit{QuickNetLarge} & $22.8M$ & $\left ( 14, 14, 128 \right )$ & $21.3M$ & $56K$ & $99.7\%$ & \\
 \bottomrule
\end{tabular}
\caption{The table represents a comparison of memory usage (\# parameters) for different BNN models. With $B$ we report the number of binary weights that can be updated during back-propagation; with $NB$ the number of non-binary weights. The choice of latent replay (LR) level is discussed in Section \ref{section_experiments}. It is worth noting that the largest part of memory weights is used by binary weights.}
\label{Tab:memory_utilization_different_bitwidths}
\end{table*}

\subsection{Quantized Backpropagation}
\label{quantized_back_prop}

Drawing upon the findings presented in the works of Gupta et al.~\cite{gupta2015deep}, Das et al.~\cite{das2018mixed}, and Banner et al.~\cite{banner2018scalable}, it is evident that the quantization of gradients stands out as the primary contributor to accuracy degradation during the training process. Therefore, we advocate for a quantization scheme akin to that introduced in \cite{vorabbi2023device}. In this scheme, we employ two distinct sets of quantization bits for the forward and backward passes.

The back-propagation algorithm operates in an iterative manner to calculate the gradients of the loss function (denoted as$\mathcal{L}$) with respect to the input $a^{l-1}$ for the layer $l$:

\begin{equation}
    g^{l} = \frac{\partial \mathcal{L}}{\partial a^{l}}
\end{equation}

starting from the last layer. Every layer in the network is tasked with computing two sets of gradients to execute the iterative update process. The first set corresponds to the layer activation gradient w.r.t. the inputs, which serves the purpose of propagating gradients backward to the previous layer. Considering a linear layer, where $a^{l}=W^{l}\cdot a^{l-1}$ and $\frac{\partial a^{l}}{\partial a^{l-1}}=W^{l}$, the gradients can be computed as follows:

\begin{equation}
\label{eq_grad_activations}
    g^{l-1} = \frac{\partial \mathcal{L}}{\partial a^{l}} \cdot \frac{\partial a^{l}}{\partial a^{l-1}} = W^{l}g^{l}
\end{equation}

The other set is used to update the weights of layer index $l$:

\begin{equation}
\label{eq_grad_weights}
    g_{w}^{l} = \frac{\partial \mathcal{L}}{\partial a^{l}} \cdot \frac{\partial a^{l}}{\partial W^{l}} = a^{l-1}g^{l}
\end{equation}

Based on Eq. \ref{eq_grad_activations} and \ref{eq_grad_weights}, the backward pass requires approximately twice Multiply-And-Accumulate (MAC) operations compared to the forward pass and therefore the gradient quantization becomes essential to efficiently train neural network models on-device.
The quantization of weights and gradients (Eq. \ref{eq_grad_activations} and \ref{eq_grad_weights}) is implemented through Eq. \ref{quant_formula} and can be visually summarized in Fig. \ref{fig:quantization_scheme}; as shown in \cite{banner2018scalable, vorabbi2023device}, backward pass usually needs higher bitwidth to preserve the directionality of the weight tensor and, based on that, we propose to use lower bitwidth during forward pass (Fig. \ref{fig:quantization_scheme}, $q_{f}$ bits, green path) to minimize latency and more bits for the backward pass to be more accurate in gradient representation (Fig. \ref{fig:quantization_scheme}, $q_{b}$ bits, purple path).
Considering the constrained memory resources available on embedded devices, accurately estimating the memory requirements of the learning algorithm becomes imperative. We can categorize memory into two distinct types: the memory utilized by the CL method (\textit{e.g.} the replay memory) and the memory necessary to store intermediate tensors during the forward pass, which are subsequently used in the backpropagation, along with the model weights. In this context, we will focus mainly on the latter aspect, particularly for binary layers where $q_{f}$ is fixed at 1-bit while $q_{b}$ can vary depending on the desired level of accuracy. In Table \ref{Tab:memory_utilization_different_bitwidths}, we present an assessment of the memory usage for representing binary weights of trainable layers on-device. It is worth noting that binary weights, as indicated in the fifth column of the same table, constitute a substantial portion of the total model parameters. Consequently, reducing $q_{b}$ to 1-bit offers significant memory savings in comparison to a more conventional approach where $q_{b}$ is set to $16$ bits. The reduction in memory usage exhibits an almost linear relationship with the number of bits utilized. 
We distinguish $q_{b}$ between binary and non-binary layers to apply different quantization bitwidts, as elaborated in Section \ref{section_experiments}, which demonstrates that it is feasible to maintain accuracy while significantly reducing $q_{b}$ for binary layers. Denoting $q_{b}^{bin}$ and $q_{b}^{non-bin}$ as the quantization settings for binary and non-binary layers, respectively, in Section \ref{section_experiments} we illustrate that setting $q_{b}^{bin}$ to 1-bit results in minimal accuracy loss compared to higher quantization bitwidths.

\section{Experiments}
\label{section_experiments}

We evaluate our methods on three classification datasets: CORe50\cite{lomonaco2017core50}, CIFAR10 \cite{krizhevsky2009learning} and CIFAR100\cite{krizhevsky2009learning} with different BNN architectures. The BNN models employed for CORe50 have been pre-trained on ImageNet through the Larq repository\footnote{\url{https://docs.larq.dev/zoo/api/sota/}}; differently, the models used for CIFAR10 and CIFAR100 have been pre-trained on TinyImageNet\cite{le2015tiny}. For each dataset, we conducted several tests using a different number of quantization bits (both for forward and backward passes) with the same training procedure. In addition to the work of Vorabbi et al.\cite{vorabbi2023device}, in our experiments we kept different bitwidths for binary and non-binary layers because, as reported in Table \ref{Tab:memory_utilization_different_bitwidths}, memory of trainable binary weights is predominant.

Hereafter we report some details about the dataset benchmarked and related CL protocols:

\begin{description}
\item[CORe50] \cite{lomonaco2017core50} It is a dataset specifically designed for Continuous Object Recognition containing a collection of $50$ domestic objects belonging to $10$ categories. The dataset has been collected in $11$ distinct sessions ($8$ indoor and $3$ outdoor) characterized by different backgrounds and lighting. For the continuous learning scenarios (NI, NC) we use the same test set composed of sessions $\#3$, $\#7$ and $\#10$; the accuracy on test set is measured after each learning experience. The remaining $8$ sessions are split in batches and provided sequentially during training obtaining $9$ experiences for NC scenario and $8$ for NI. No augmentation procedure has been implemented since the dataset already contains enough variability in terms of rotations, flips and brightness variation. The input RGB image is standardized and rescaled to the size of $128 \times 128 \times 3$.
\item[CIFAR10 and CIFAR100] \cite{krizhevsky2009learning} Due to the lower number of classes, the NC scenario for CIFAR10 contains $5$ experiences (adding $2$ classes for each experience) while $10$ are used for CIFAR100. For both datasets the NI scenario is composed by $10$ experiences. Similar to CORe50, the test set does not change over the experiences. The RGB images are scaled to the interval $\left [-1.0\: ;+1.0  \right ]$ and the following data augmentation was used: zero padding of $4$ pixels for each size, a random $32 \times 32$ crop and a random horizontal flip. No augmentation is used at test time. 
\end{description}

On CORe50 dataset, we evaluated the three binary models reported below:

\begin{description}
\item[Quicknet and QuicknetLarge] \cite{bannink2021larq} \hspace{0.05\textwidth} This network follows the previous works \cite{liu2018bi, bethge2019back, martinez2020training} proposing a sequence of blocks, each one with a different number of binary $3 \times 3$ convolutions and residual connections over each layer. Transition blocks between each residual section halve the spatial resolution and increase the filter count. QuicknetLarge employs more blocks and feature maps to increase accuracy. For Quicknet, latent replay memory has been set to quant\textunderscore conv2d\textunderscore$16$ layer by storing 1-bit activations; for QuicknetLarge the latent replay level is quant\textunderscore conv2d\textunderscore$30$. At this level (both for Quicknet and QuicknetLarge) activation has a dimensionality of $\left(7,7,256\right)$ and storing in the replay memory 1-bit activations leads to a considerable memory saving.
\end{description}

In contrast to the findings presented in \cite{vorabbi2023device}, our study did not include the \textit{Realtobinary} \cite{martinez2020training} model, as it achieved notably lower accuracy levels that were not aligned with our research objectives and goals.

For CIFAR10 and CIFAR100 datasets, whose input resolution is $32\times32$, we evaluated the following networks (pre-trained on Tiny Imagenet):

\begin{description}
\item[BiRealNet]\cite{liu2018bi} \hspace{0.05\textwidth} It is a modified version of classical ResNet that proposes to preserve the real activations before the sign function to increase the representational capability of the 1-bit CNN, through a simple shortcut. Bi-RealNet adopts a tight approximation to the derivative of the non-differentiable sign function with respect to activation and a magnitude-aware gradient to update weight parameters. We used the instance of the network that uses \textit{18-layers}\footnote{Refer to the following \url{https://github.com/liuzechun/Bi-Real-net} repository for all the details.}. The latent replay layer has been set to add\textunderscore$12$. At this level activation has a dimensionality of  $\left(4,4,512\right)$.
\item[ReactNet]\cite{liu2020reactnet} \hspace{0.05\textwidth} To further compress compact networks, this model constructs a baseline based on MobileNetV1 \cite{howard2017mobilenets} and adds a shortcut to bypass every 1-bit convolutional layer that has the same number of input and output channels. The $3 \times 3$ depth-wise and the $1 \times 1$ point-wise convolutional blocks of MobileNet are replaced by the $3 \times 3$ and $1 \times 1$ vanilla convolutions in parallel with shortcuts in React Net\footnote{Refer to the following \url{https://github.com/liuzechun/ReActNet} repository for all the details.}. As for Bi-Real Net, we tested the version of React Net that uses  \textit{18-layers}. The latent replay layer has been set to add\textunderscore$12$ layer. At this level activation has a dimensionality of  $\left(4,4,512\right)$.
\end{description}

In our experimental setup, we discovered that reducing the number of epochs in each learning experience had minimal impact on model accuracy. Consequently, we empirically set the number of epochs to $5$, thus constraining the training time on-device platform. Across all classification tasks, we utilized the Cross Entropy loss function in conjunction with Stochastic Gradient Descent (SGD) as the optimizer. The former was chosen due to its simplicity in derivative computation when combined with the Softmax activation function. The latter was preferred for its computational efficiency, offering lower overhead compared to more complex algorithms like Adam~\cite{kingma2014adam}. In our experiments, the ratio of $B_{N}$ to the batch size of the latent activations sampled from the replay memory is set at $1/4$. Both weight and activation binarization were performed during training, including both the first training experience and on-device stages. This choice requires the implementation of a quantized backward pass technique for all the non-differentiable functions, specifically the binarization functions (using Eq. \ref{eq_grad_activations} and \ref{quant_formula}). To assess model accuracy during on-device training, we developed the quantized backward steps for all layers employed by the previously described models.

Our experiments primarily concentrated on the NC scenario. As highlighted in \cite{pellegrini2020latent}, the adoption of a latent replay memory did not significantly enhance model accuracy in the NI context. Moreover, the NC scenario more closely resembles real-world applications where the model's recognition capability must be expanded to accommodate new, previously unknown classes.

\subsection{Accuracy comparison}
\label{section_accuracy_result}

\begin{figure}[]
\centering
\includegraphics[scale=0.42]{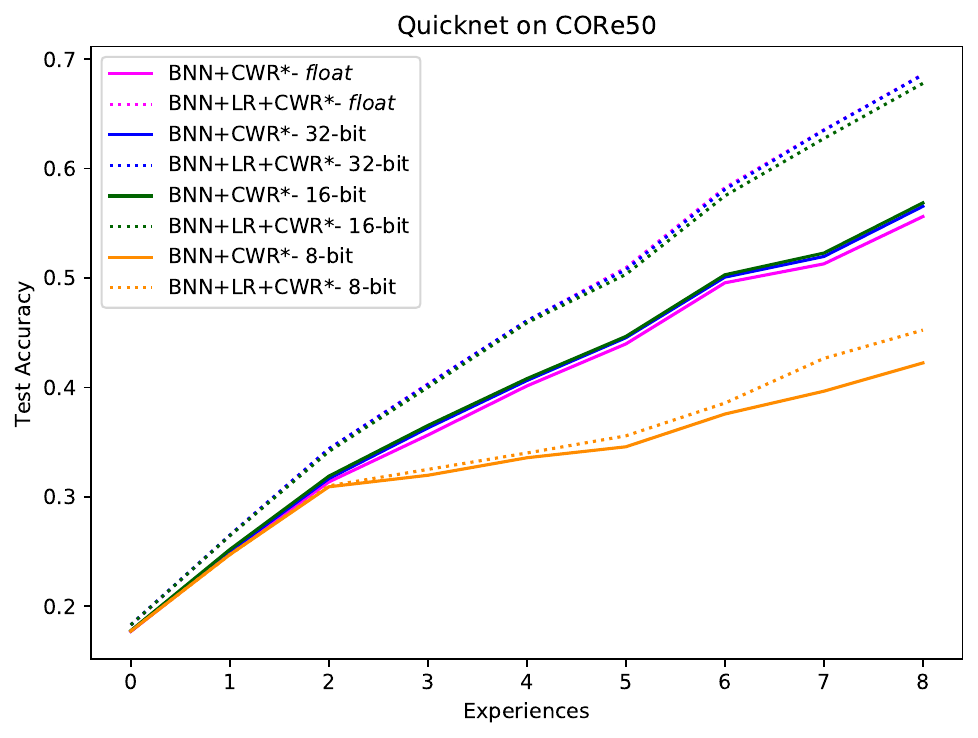}
  	\caption{Accuracy comparison of our solution (BNN+LR+CWR*) with previous work BNN+CWR*~\cite{vorabbi2023device} on CORe50 using \textit{quick} model.}
  	\label{fig:comparison_previous_cwr_quicknet_core50}
\end{figure}

\begin{figure}[]
\centering
\includegraphics[scale=0.42]{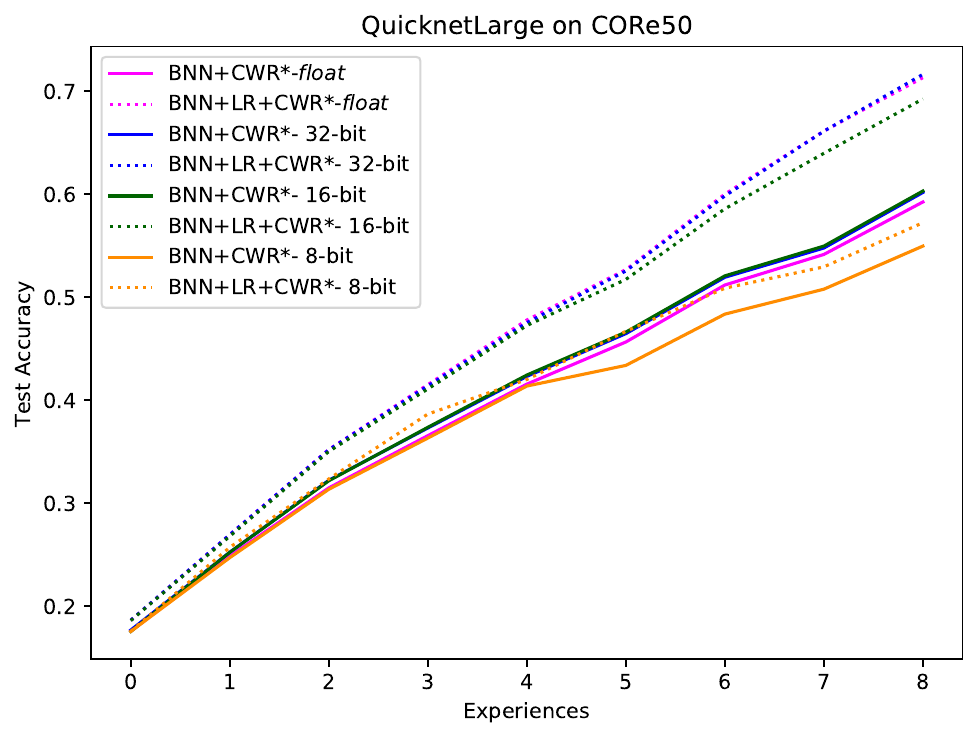}
  	\caption{Accuracy comparison of our solution (BNN+LR+CWR*) with previous work BNN+CWR*~\cite{vorabbi2023device} on CORe50 using \textit{QuickNetLarge} model.}
  	\label{fig:comparison_previous_cwr_with_ours_quicknetlarge_core50}
\end{figure}

\begin{figure}[]
\centering
\includegraphics[scale=0.42]{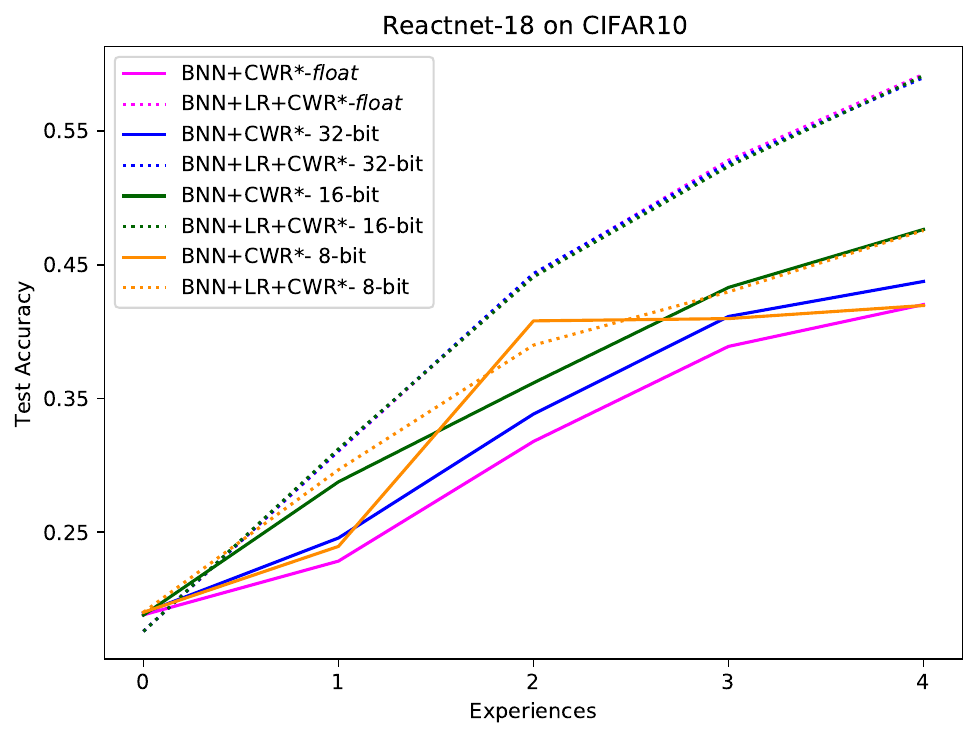}
  	\caption{Accuracy comparison of our solution (BNN+LR+CWR*) with previous work BNN+CWR*~\cite{vorabbi2023device} on CIFAR10 using \textit{Reactnet} model.}
  	\label{fig:comparison_previous_cwr_with_ours_reactnet_cifar10}
\end{figure}

\begin{figure}[]
\centering
\includegraphics[scale=0.42]{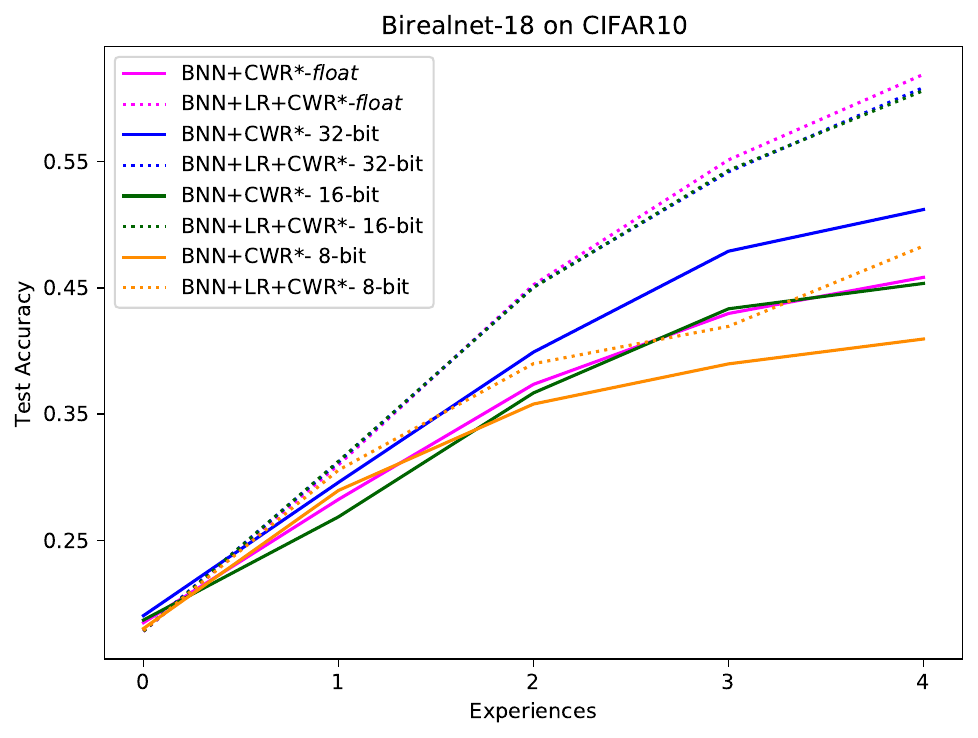}
  	\caption{Accuracy comparison of our solution (BNN+LR+CWR*) with previous work BNN+CWR*~\cite{vorabbi2023device} on CIFAR10 using \textit{Birealnet} model.}
  	\label{fig:comparison_previous_cwr_with_ours_birealnet_cifar10}
\end{figure}

\begin{figure}[]
\centering
\includegraphics[scale=0.42]{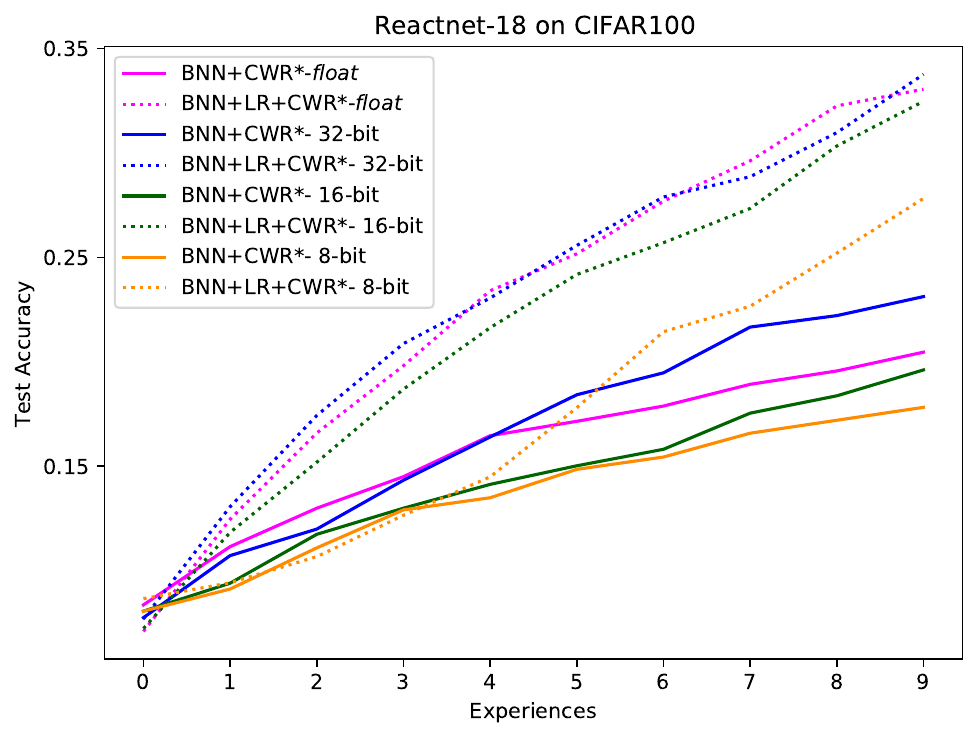}
  	\caption{Accuracy comparison of our solution (BNN+LR+CWR*) with previous work BNN+CWR*~\cite{vorabbi2023device} on CIFAR100 using \textit{Reactnet} model.}
  	\label{fig:comparison_previous_cwr_with_ours_reactnet_cifar100}
\end{figure}

\begin{figure}[]
\centering
\includegraphics[scale=0.42]{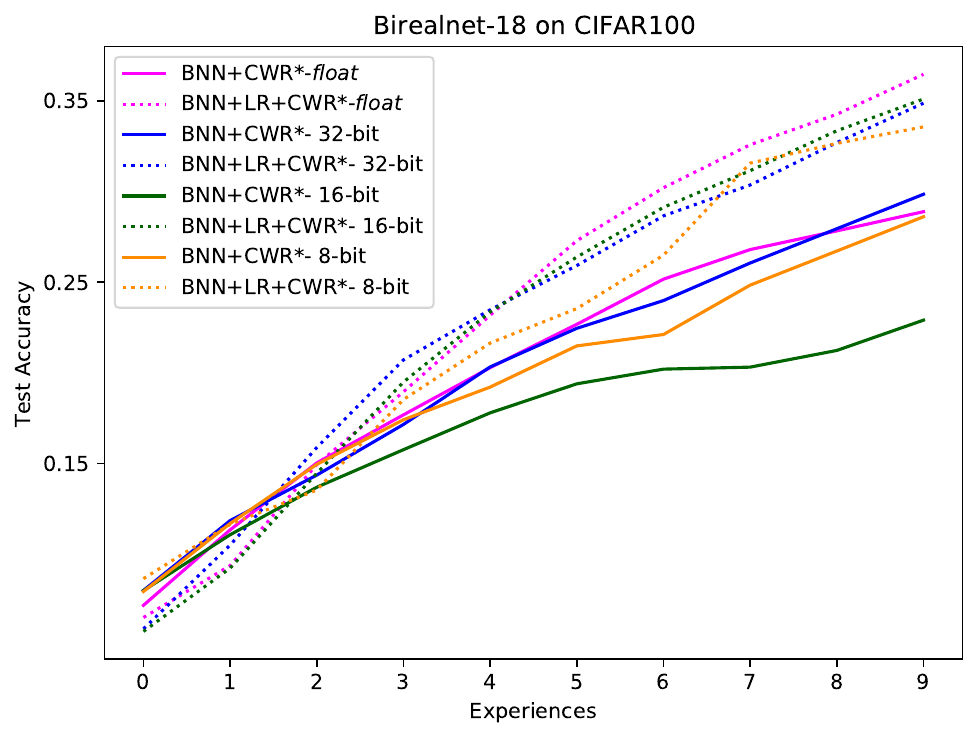}
  	\caption{Accuracy comparison of our solution (BNN+LR+CWR*) with previous work BNN+CWR*~\cite{vorabbi2023device} on CIFAR100 using \textit{Birealnet} model.}
  	\label{fig:comparison_previous_cwr_with_ours_birealnet_cifar100}
\end{figure}

To assess the accuracy of our proposed solution, we initiated our evaluation by comparing it with prior work, specifically BNN+CWR*~\cite{vorabbi2023device}, where only the final classification layer is trained on-device, without employing a replay memory. We conducted a series of tests with varying quantization bitwidths for both forward and backward passes. In Fig. \ref{fig:comparison_previous_cwr_quicknet_core50}, \ref{fig:comparison_previous_cwr_with_ours_quicknetlarge_core50}, \ref{fig:comparison_previous_cwr_with_ours_reactnet_cifar10}, \ref{fig:comparison_previous_cwr_with_ours_birealnet_cifar10}, \ref{fig:comparison_previous_cwr_with_ours_reactnet_cifar100} and \ref{fig:comparison_previous_cwr_with_ours_birealnet_cifar100} we present accuracy comparisons between BNN+CWR* with the current method, denoted as \textbf{BNN+LR+CWR*}, across different datasets: CORe50, CIFAR10 and CIFAR100. Each figure illustrates the performance improvement of the new method for all quantization settings tested, encompassing floating-point arithmetic, 32-bit, 16-bit and 8-bit quantized representations. It is noteworthy that, in this assessment, we applied the same quantization bitwidths ($q_{b}$) for both binary ($q_{b}^{bin}$) and non-binary ($q_{b}^{non-bin}$) layers during the backward pass, as BNN+CWR* does not distinguish these cases. The results consistently demonstrate that our BNN+LR+CWR* approach outperforms previous results, not only when using floating-point arithmetic but also for quantized implementations. This underscores the superior performance achieved by BNN+LR+CWR*. In our solution, we observed that employing $q_{b}=8$ in BNN+LR+CWR* leads to a notable drop in accuracy compared to higher quantization bitwidth settings, aligning with the outcomes obtained by BNN+CWR*. This reaffirms the importance of using higher bitwidth representations during the backward pass to preserve model accuracy. For the experiments, we utilized $LR_{size}=1500$ for CORe50, $LR_{size}=300$ for CIFAR10 and $LR_{size}=3000$ for CIFAR100 as our replay memory sizes.

\subsection{Reducing Storage in Latent Replay} \label{reduction_memory}

\begin{figure}[!h]
\centering
\begin{subfigure}{\linewidth}
\centering
    \includegraphics[scale=0.46]{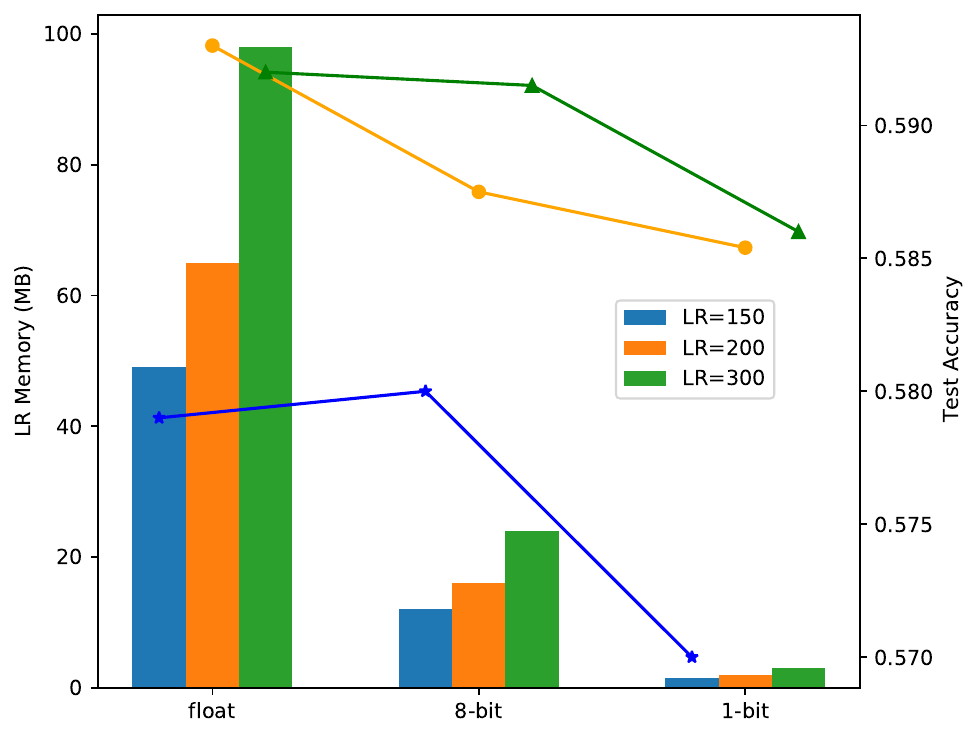}
    \caption{Reactnet-18 on CIFAR10.}
    \label{fig:LR_memory_size_comparison_cifar10}
\end{subfigure}
\vfill
\bigskip
\begin{subfigure}{\linewidth}
\centering
    \includegraphics[scale=0.46]{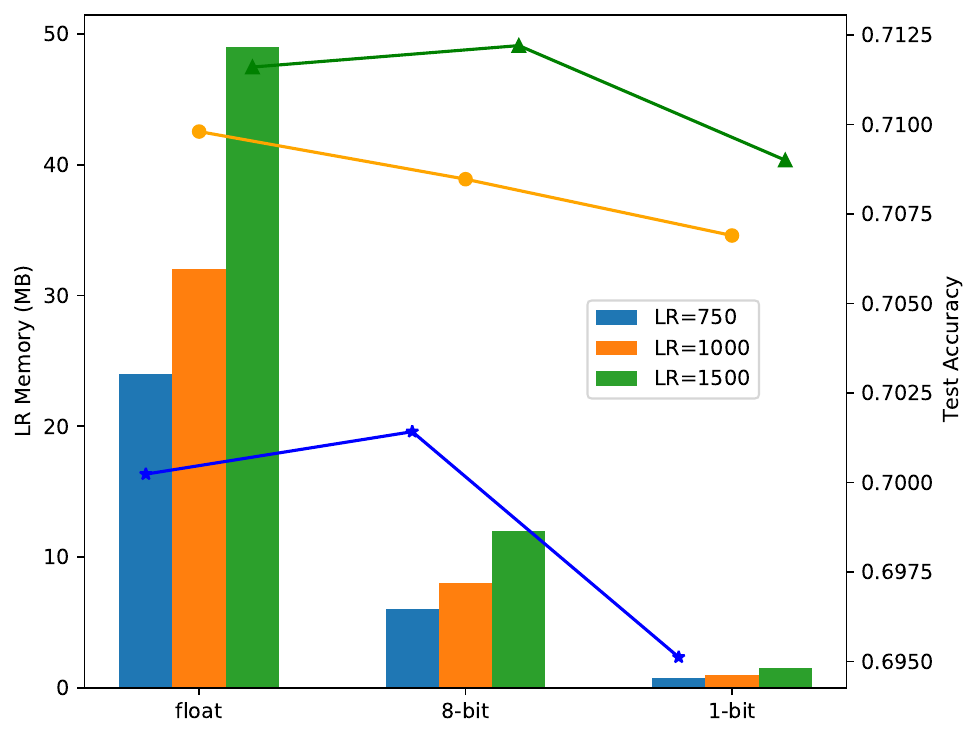}
    \caption{Quicknet on CORe50.}
    \label{fig:LR_memory_size_comparison_core50}
\end{subfigure}
\caption{LR memory requirement using different quantization levels and corresponding test set accuracy on CIFAR10 (\subref{fig:LR_memory_size_comparison_cifar10}) and CORe50 (\subref{fig:LR_memory_size_comparison_core50}). We considered $15, 20$ and $30$ elements for each class inside LR; for case (\subref{fig:LR_memory_size_comparison_cifar10}) we adopted Reactnet-18 model while in (\subref{fig:LR_memory_size_comparison_core50}) we used Quicknet.}
\label{fig:LR_memory_size_comparison}
\end{figure}

The storage requirements of the latent replay memory are closely interlinked with the bitwidths utilized to represent latent activations. As the bitwidths increase, so does the memory footprint of LR. In our approach we capitalize on the 1-bit activations inherent to BNNs to significantly mitigate the need for high-memory storage while maintaining a minimal accuracy gap, as depicted in Fig. \ref{fig:LR_memory_size_comparison}. Our experiments demonstrate that BNN models can attain a minimal accuracy gap on both CIFAR10 and CORe50 datasets, even when adopting 1-bit latent activations for LR. This translates to a huge memory reduction of $32 \times$ when compared to using floating-point latent activations. 
In our analysis, we considered various sizes for the LR memory, with $15, 20$ and $30$ elements allocated for each class. Importantly, we observed that the number of past samples in LR had a relatively minor impact on model accuracy, with the accuracy loss being within $1 \%$. Utilizing 1-bit latent activations for LR opens the possibility to scale up applications to accommodate thousands of classes, as illustrated in Fig. \ref{fig:LR_memory_size_comparison}, thanks to the substantial reduction in memory constraints achieved.

\subsection{Splitting $q_{b}$ in $q_{b}^{bin}$ and $q_{b}^{non-bin}$}
\label{section_splitting_q2_for_binary}

\begin{figure}[!h]
\centering
\begin{subfigure}{\linewidth}
\centering
    \includegraphics[scale=0.46]{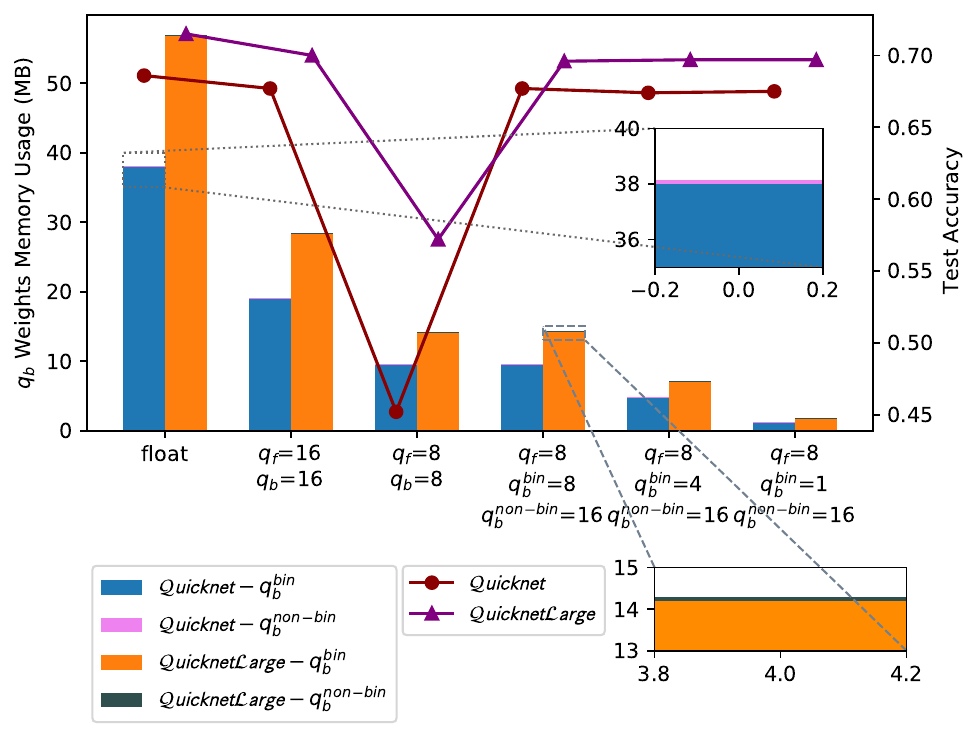}
    \caption{Quicknet and QuicknetLarge on CORe50.}
    \label{fig:comparison_bits_widths_accuracy_core50}
\end{subfigure}
\vfill
\bigskip
\begin{subfigure}{\linewidth}
\centering
    \includegraphics[scale=0.46]{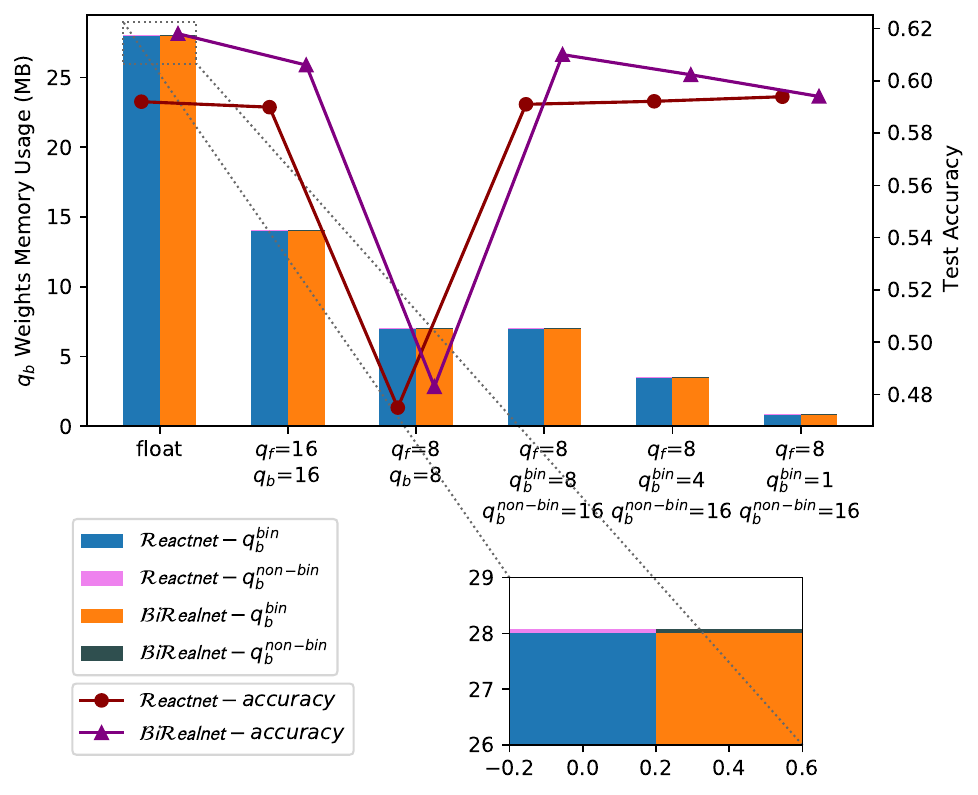}
    \caption{Reactnet-18 on CIFAR10.}
    \label{fig:comparison_bits_widths_accuracy_cifar10}
\end{subfigure}
\caption{$q_{b}$ memory requirement using different quantization bitwidths for backward layer on CORe50 (\subref{fig:comparison_bits_widths_accuracy_core50}) and CIFAR10 (\subref{fig:comparison_bits_widths_accuracy_cifar10}).}
\label{fig:comparison_bitwidths_accuracy}
\end{figure}

As highlighted in Table \ref{Tab:memory_utilization_different_bitwidths}, the memory footprint of BNN weights is predominantly occupied by trainable binary weights, encompassing nearly $100\%$ of the memory. Conventionally, a binary layer is trained using latent floating-point weights \cite{helwegen2019latent}. However, if we were to replicate this approach on the device, it would result in a substantial increase in memory storage requirements during backpropagation stage, as it would require setting $q_{b}^{bin}=32-bit$. 
The quantization methodology proposed in Section \ref{quantized_back_prop} offers a potential solution to mitigate this constraint by reducing $q_{b}$ to 8 bits. However, as depicted in Figure \ref{fig:comparison_bits_widths_accuracy_core50} and \ref{fig:comparison_bits_widths_accuracy_cifar10}, such a reduction in bitwidths would lead to a noticeable accuracy drop in the model.
To address this challenge, we evaluated the impact of distinct quantization levels for binary weights ($q_{b}^{bin}$) and non-binary weights ($q_{b}^{non-bin}$). Specifically, we experimented with representing $q_{b}^{bin}$ using both 4 bits and 1 bit. Our findings, as shown in Figure \ref{fig:comparison_bitwidths_accuracy}, indicate that 4-bit representation for binary layers does not introduce a substantial accuracy loss. Moreover, employing a 1-bit representation of weights during the back-propagation stage is feasible, as binary weights remain frozen during on-device learning. In this scenario, the model still effectively preserves accuracy. This latest result carries significant implications for on-device learning, as it simplifies the computational burden by requiring backward steps only for non-binary layers, primarily those employing $q_{b}^{non-bin}=16-bits$, as observed in our experiments.

\subsection{Efficiency Evaluation}
\label{eficiency_estimation}
To demonstrate the applicability of our approach on real-world embedded boards, we provide an estimation analysis of the on-device performance. 
For this evaluation, we select two popular boards commonly used in the IoT paradigm, both based on the single-thread ARMv8 platform: Raspberry Pi 3B and Raspberry Pi 4B. 
Based on the efficiency analysis reported in \cite{bannink2021larq, pellegrini2020latent}, we report in Table \ref{efficiency_table} the inference and backward timings of our BNN+LR+CWR* method compared to a non-binary solution (using a Mobilenetv2) \cite{pellegrini2020latent}: the results obtained adopting Mobilenetv2 rely on floating-point precision for layers from LR up to the classification head. The frozen backbone is quantized using 8-bit (latent activations are stored with 8-bit precision) and executed with Tensorflow-Lite. Instead, BNN+LR+CWR* employs Quicknet model with the following quantization setting: $q_{f}=8, q_{b}^{bin}=8, q_{b}^{non-bin}=16$; the framework used to execute binary inference is LCE~\cite{bannink2021larq}. The image input size considered is $224 \times 224$ and the batch size is $1$. Our empirical evaluation for backward pass shows that our BNN+LR+CWR* can achieve a minimum speedup of $2 \times$ compared to a non-binary solution. In our evaluation we consider the worst-case scenario for backward step by setting $q_{b}^{bin} = 8$; instead, by setting $q_{b}^{bin} = 1$, the speedup reported in the fifth column of Table \ref{efficiency_table} should improve significantly.

\begin{table*}[]
\centering
\resizebox{1\linewidth}{!}{
\begin{tabular}{rcc|c|cc|cc|c}
\toprule
\multicolumn{1}{c}{\textbf{Model}}       & \multicolumn{2}{c|}{\textbf{Raspberry}} & \textbf{Binary} & \multicolumn{2}{c|}{\textbf{Quantization}} & \textbf{Forward} & \textbf{Backward} & \textbf{Speedup} \\
            & 3B            & 4B            &        & $q_{f}$             & $q_{b}$              &   \multicolumn{2}{c|}{(ms)}                &         \\
\midrule
Mobilenetv2~\cite{howard2017mobilenets} & \ding{51}             &               &        & 8-bit          & float           & $340$     & $134$      & $1.0 \times$      \\
Quicknet~\cite{bannink2021larq}   & \ding{51}             &               & \ding{51}      & 1-bit          & 16-bit          & $\mathbf{160}$     & $\mathbf{55}$       & $\mathbf{2.0 \times}$      \\
\midrule
Mobilenetv2~\cite{howard2017mobilenets} &               & \ding{51}             &        & 8-bit          & float           & $225$     & $90$       & $1.0 \times$      \\
Quicknet~\cite{bannink2021larq}    &               & \ding{51}             & \ding{51}      & 1-bit          & 16-bit          & $\mathbf{105}$     & $\mathbf{38}$       & $\mathbf{2.2 \times}$    \\
\bottomrule
\end{tabular}
}
\caption{Efficiency comparison of our method implemented on two different embedded boards, \textit{i.e.} Raspberry Pi 3B and 4B, using Mobilenetv2 and Quicknet model. As shown, our solution achieves up to $2.2\times$ speedup on the same platform.}
\label{efficiency_table}
\end{table*}

\section{Conclusion}
On-device training holds great potential in the realm of the IoT, as it can facilitate the widespread adoption of deep learning solutions. In this study, our primary focus was the implementation of Binary Neural Networks (BNNs) in combination with Continual Learning algorithms, an approach not yet fully investigated in the literature.
In particular, we propose the use of the CWR* method with the support of a replay memory, implementing several customized quantization schemes tailored to alleviate memory constraints and computational bottlenecks during the back-propagation stage.
% BNN+LR+CWR* on edge devices, harnessing the remarkable computational efficiency provided by binary neural networks.  
Summarizing, experimental achievements in this work include the following:

\begin{itemize}
    
    \item \textbf{Reduced memory usage}: we significantly reduced the memory storage required for replay memory by employing 1-bit latent activations, as opposed to the state-of-the-art approach that employs 8-bit precision. A limited storage requirement is a key element in addressing on-device training, especially with embedded systems with a limited storage capability.
    
    \item \textbf{Improved model accuracy}: we improve the accuracy obtained across different binarized backbones and the BNN+CWR* approach. Specifically, we reduce the gap in performance that commonly affects BNNs by introducing a latent replay approach as a safeguard against catastrophic forgetting.
    
    \item \textbf{Efficiency in backpropagation}: we minimize the computational effort related to the backpropagation of the latent replay through a proper quantization scheme. In this manner, we combine the good performance of the model with limited computation requirements for the learning phase. This achievement, in combination with reduced memory usage, paves the way for future on-device and real-world training of learning systems.
\end{itemize}

A variety of future work can be planned based on the technological advancements introduced in this paper. For instance, we plan to effectively implement and optimize the approach proposed in this paper for the specific ARM CPUs, a popular family of processors often used in IoT devices.
In addition, we envisage the possibility of exploiting their instruction set, including NEON extensions, to further optimize the proposed method in terms of computational load and efficiency.

%\bibliographystyle{apalike}
%{\small
%\bibliography{example}}
\printbibliography

\end{document}